\documentclass{article}
\usepackage{spconf,amsmath,graphicx,hyperref}
\usepackage{amsfonts}
\usepackage{booktabs}
\usepackage[table]{xcolor} 
\usepackage{multirow}

\usepackage{listings}
\usepackage{xcolor}
\usepackage{amsmath,amssymb}

\title{{\rm REC-RL}: Referring Expression Counting via Gaussian and Range-Based Reward Optimization}

\name{Hui Liu\textsuperscript{\dag}, Yunlai Teng\textsuperscript{\dag}, Kunlong Bai, Pengfei Qi, Haotian Yan\textsuperscript{*}, Liang Li, Junlan Feng\thanks{\textsuperscript{\dag} These authors contributed equally. }\thanks{ \textsuperscript{*} Work done during an internship at JIUTIAN Research Institute.}}
\address{JIUTIAN Research, Beijing, China}

\begin{document}
%
\maketitle
\begin{abstract}
Referring expression counting (REC) is an intention-driven task that requires context-aware visual reasoning. While recent vision–language models incorporate language for visual understanding, most existing REC methods rely on rule-based reinforcement learning with rewards focused primarily on final accuracy, overlooking the quality of intermediate reasoning. We propose REC-RL, a reinforcement learning framework that introduces a \textbf{think–range–answer} paradigm to explicitly optimize the visual reasoning process. REC-RL employs Group Relative Policy Optimization and two lightweight rewards: an accuracy reward that combines range-based interval supervision with Gaussian-based precision guidance, and a format reward that enforces structured outputs. By modeling intermediate focus prediction as internal decision-making, REC-RL avoids additional annotations and better aligns with human perception. Extensive experiments demonstrate consistent improvements over strong baselines and robust generalization across benchmarks.
\end{abstract}
\begin{keywords}
Reinforcement Learning, Vision-Language Models, Referring Expression Counting
\end{keywords}
\section{Introduction}
\label{sec:intro}
Referring Expression Counting (REC) is a fine-grained computer vision task that aims to quantify objects specified by both category and contextual attributes \cite{dai2024referring}. Unlike conventional class-level counting, REC requires understanding compositional queries such as ``green pears on the table'' within a broader category like ``fruit,'' where attributes and spatial context jointly define the target set. As a result, REC relies heavily on visual grounding, which localizes image regions conditioned on language. Most existing REC systems are built upon vision-language models (VLMs) trained via supervised fine-tuning (SFT). However, SFT suffers from two major limitations: it requires large-scale annotated datasets, and open-source SFT models often generalize poorly to out-of-domain scenarios, limiting the practical applicability of REC \cite{alayrac2022flamingo}.

Rule-based reinforcement learning (RL) has recently emerged as an efficient alternative to SFT for optimizing VLMs. By leveraging predefined reward functions, RL enables effective fine-tuning with only dozens to thousands of samples, significantly reducing annotation costs. Recent progress highlights its potential: DeepSeek R1 \cite{guo2025deepseek} demonstrates that simple rule-based rewards can induce reasoning capabilities in large language models, while Group Relative Policy Optimization (GRPO) \cite{shao2024deepseekmath,yu2025dapo} further stabilizes learning for long Chain-of-Thought (CoT) reasoning \cite{wei2022chain}. Beyond text-only settings, multimodal extensions such as R1-OneVision \cite{yang2025r1} and VLM-R1 \cite{shen2025vlmr1} successfully adapt this paradigm to vision-language models, where rule-based reward functions provide reliable outcome supervision \cite{deng2025openvlthinker}. Collectively, these studies suggest that RL is particularly effective for deterministic tasks like REC, as it yields stable and interpretable training signals.

Despite this progress, existing REC methods still exhibit critical limitations. Traditional approaches such as DINO-REC \cite{dai2024referring} lack comprehensive understanding of complex, compositional instructions. More broadly, current RL frameworks for VLMs \cite{yang2025r1,shen2025vlmr1,Liu_2025_ICCV} rely almost exclusively on binary outcome rewards to guide learning \cite{deng2025openvlthinker}. Such sparse supervision fails to enforce high-quality reasoning and often encourages shortcut behaviors that bypass meaningful visual analysis. While these shortcuts may suffice for simple cases, they are inadequate for complex REC queries that require deep visual understanding and multi-step reasoning \cite{lightman2023let}.

To address these limitations, we propose \textbf{REC-RL}, a reinforcement learning framework tailored for referring expression counting via \textbf{Gaussian and range-based reward optimization}. Instead of relying on sparse outcome rewards, REC-RL introduces a structural shift from outcome-only optimization to a \textbf{think-range-answer} paradigm. This paradigm fosters verifiable reasoning and effectively mitigates shortcut behaviors. Crucially, this shift represents more than a marginal refinement: as demonstrated in Table 2, while the Gaussian-guided reward ($\alpha=0$) already marginally outperforms the baseline by providing a non-linear feedback signal that more appropriately reflects the degree of prediction accuracy , the integration of range-based auxiliary rewards leads to a significant performance leap. By optimizing these components via the GRPO algorithm \cite{shao2024deepseekmath}, our model achieves state-of-the-art results on the REC task using only 2,000 training samples (Table 1). This proves that our paradigm is uniquely effective at resolving the core challenges of referring expression counting with high data efficiency.

Experimental results demonstrate that REC-RL substantially improves visual understanding in vision-language models compared to supervised fine-tuning. On the REC-8K benchmark \cite{dai2024referring}, REC-RL achieves strong performance and exhibits improved generalization on complex real-world datasets.

\begin{figure*}
	\centering
	\includegraphics[width=\linewidth]{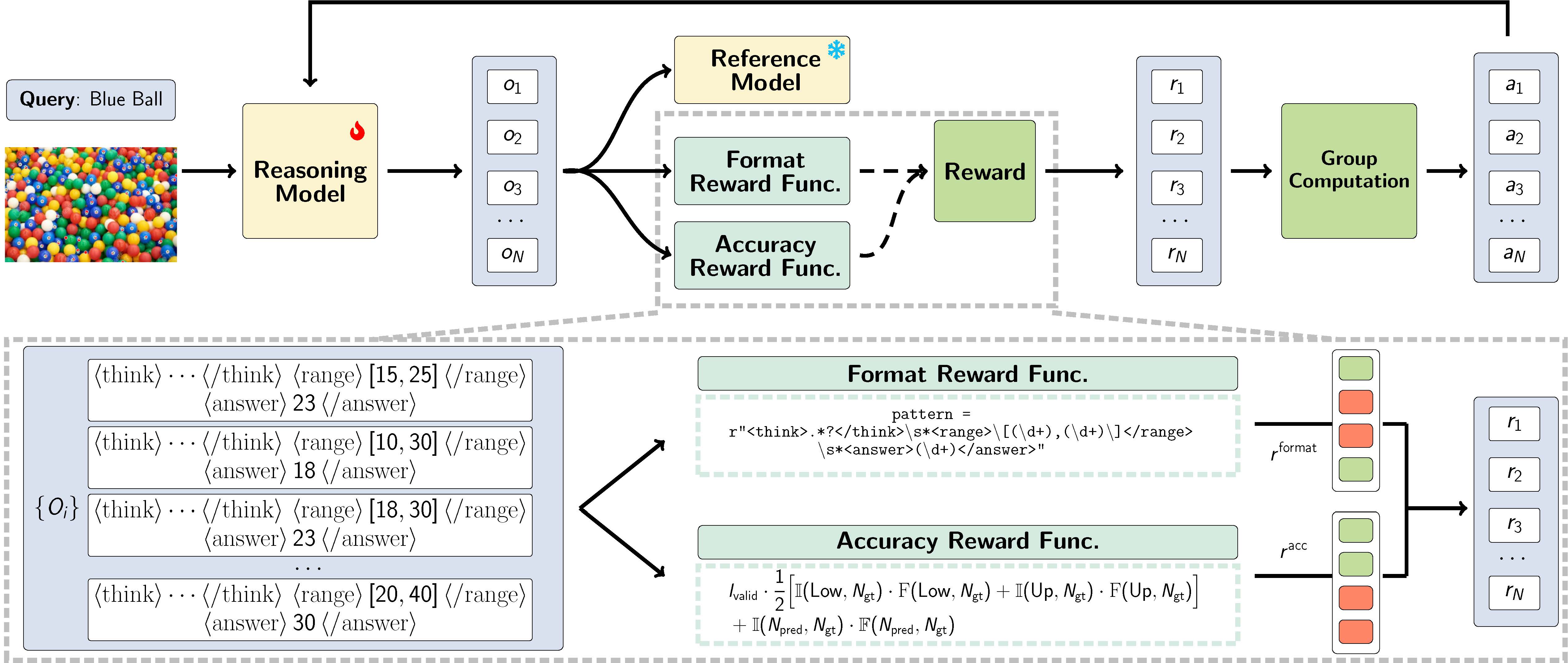}
	\caption{Framework of REC-RL. In REC-RL, GRPO enhances reasoning capabilities further, enabling REC-RL to achieve superior generalization by pushing the model’s reasoning limits. Specifically, reward evaluation
		consists of format reward and accuracy reward. For brevity, some text instructions are omitted.}
	\label{fig:pipeline}
\end{figure*}

\section{Method}
\label{sec:format}

The REC-RL proposes a novel framework tailored for referring expression counting via \textbf{Gaussian and Range-Based Reward Optimization}. As illustrated in Fig.\ref{fig:pipeline}, for a given question $q$, the GRPO algorithm first samples $N$ candidate responses $\left\lbrace o_{1}, o_{2}, \cdots, o_{N} \right\rbrace $ from the policy model $\theta_{\text{old}}$, where each response is structured following the think–range–answer process. Subsequently, each response $o_{i}$ is evaluated using a rule-based verifiable reward function, which comprises two core components: accuracy reward and format reward.

\subsection{Reward Functions}
\label{ssec:subhead}
\subsubsection{Accuracy reward ($r^{acc}$):}
The accuracy reward assesses the model’s object count prediction performance, with two sub-components: (1) \textit{range-based reward} $r^{range}$, which evaluates the rationality and precision of the predicted interval
 $[\text{Low}, \text{Up}]$; (2) \textit{gaussian reward} $r^{ans}$, which evaluates the precision of the predicted count $N_{\text{pred}}$.
 
To quantify these rewards, we first define two reusable functions for any predicted value $y$ and ground truth $N_{\text{gt}}$, followed by the interval validity indicator:

\textbf{Validity Indicator} (checks if $N_{\text{pred}}$ is within 50\% relative error of $N_{\text{gt}}$, where $\epsilon$ is a small constant to avoid division by zero.):
\begin{equation}
\mathbb{I}(N_{\text{pred}}, N_{\text{gt}}) = 
\begin{cases}
	1, & \text{if } \left| \dfrac{N_{\text{pred}} - N_{\text{gt}}}{\max(N_{\text{gt}}, \epsilon)} \right| < 0.5 \\
	0, & \text{otherwise}
\end{cases}
\end{equation}

\noindent \textbf{Gaussian Function.} 
To more precisely measure the discrepancy between the predicted value $N_{\text{pred}}$ and the ground truth $N_{\text{gt}}$, simple linear metrics are often insufficient for capturing fine-grained counting sensitivity. 
Therefore, we adopt a \textbf{non-linear reward function} to shape the optimization landscape with a soft, continuous signal. 
Specifically, we employ a Gaussian formulation defined as:

\begin{equation}
    \mathbb{F}(y, N_{\text{gt}}) = \exp\left(-k \cdot \left(\frac{N_{\text{pred}} - N_{\text{gt}}}{\max(N_{\text{gt}}, \epsilon)}\right)^2\right)
\end{equation}

\noindent where $k$ is a scaling hyperparameter. 
The hyperparameter $k$ is determined by the $3\sigma$ principle. To ensure the reward signal vanishes when the relative error exceeds a tolerance boundary of $0.5$, we set $3\sigma = 0.5$, yielding a theoretical coefficient of $k = 1/(2\sigma^2) = 18$. In practice, we adopt a slightly stricter penalty ($k = 20$) to sharpen the reward peak. 

We define the interval validity criterion $I_{valid}$ to ensure the predicted interval $[Low, Up]$ encapsulates the ground truth $N_{\text{gt}}$, we propose a predicted interval validity criterion, defined as follows:
\begin{equation}
I_{\text{valid}} = 
\begin{cases}
1, & \text{if } \text{Low} \leq N_{\text{gt}} \leq \text{Up} \\
0, & \text{otherwise}
\end{cases}
\end{equation}

Using the indicators defined above, we calculate the two components of the accuracy reward as follows:

\textbf{Range-Based Reward $r^{\text{range}}$.} This reward is only non-zero if the predicted interval is valid ($I_{\text{valid}} = 1$). For valid intervals, it sums the precision contributions of the lower and upper bounds, with an exponential term that sharply penalizes deviations from $N_{\text{gt}}$ (rewarding tight alignment):
\begin{align}
r^{\text{range}} = I_{\text{valid}} \cdot \frac{1}{2} \Big[
&\mathbb{I}(\text{Low}, N_{\text{gt}}) \cdot \mathbb{F}(\text{Low}, N_{\text{gt}}) \nonumber \\
+ &\mathbb{I}(\text{Up}, N_{\text{gt}}) \cdot \mathbb{F}(\text{Up}, N_{\text{gt}})
\Big].
\end{align}

When the relative error of a bound approaches 50\%, $\mathbb{F}$ decays precipitously. This mechanism dual-purpose: it heavily penalizes large deviations from $N_{\text{gt}}$ while rewarding the derivation of highly precise interval bounds.

\textbf{Answer-Based Reward.}
This reward evaluates the precision of the prediction $N_{\text{pred}}$, using the same exponential penalty for deviations from $N_{\text{gt}}$:
\begin{equation}
r^{\text{ans}} = \mathbb{I}(N_{\text{pred}}, N_{\text{gt}}) \cdot \mathbb{F}(N_{\text{pred}}, N_{\text{gt}}).
\end{equation}

\textbf{Total accuracy reward.} The final accuracy reward combines $r^{\text{range}}$ and $r^{\text{ans}}$, with a hyperparameter $\alpha$ controlling the relative weight of the range-based component:
\begin{equation}
r^{\text{acc}} = r^{\text{ans}} + \alpha \cdot r^{\text{range}}.
\end{equation}

\subsubsection{Format reward ($r^{\text{format}}$):}
The format reward enforces strict adherence to a predefined response structure. Specifically, the model is required to: (1) enclose its reasoning process within \texttt{<think>} tags, (2) report the predicted interval within \texttt{<range>} tags, and (3) output the final numerical answer within a single pair of \texttt{<answer>} tags:
\begin{lstlisting}
{<think> ... </think>
<range>[LOW, UP]</range>
<answer> ... </answer>}
\end{lstlisting}
A response that satisfies all three requirements receives $r^{\text{format}} = 1$; any deviation results in $r^{\text{format}} = 0$.

\subsection{Optimization}
The total reward for a response $o_{i}$ is defined as:
\begin{equation}
r_{i}=r^{\text{acc}}_{i}+r^{\text{format}}_{i}.
\end{equation}

To determine the relative quality of these responses, GRPO normalizes the rewards by computing their mean and standard deviation. The advantage for each response is then computed as:
\begin{equation}
A_{i}=\frac{r_{i}-mean(\left\lbrace r_1, r_2, \cdots, r_N \right\rbrace )}{std(\left\lbrace r_1, r_2, \cdots, r_N \right\rbrace )},
\end{equation}
where $A_i$ represents the advantage of the candidate response $o_{i}$ relative to the other sampled responses within the group. GRPO encourages the model to generate responses with higher advantages by updating the policy $\pi_{\theta}$ to maximize the following objective function:

\begin{equation}
\begin{aligned}
J_{GRPO}(\theta) 
&= \mathbb{E}\Bigl[ \{o_i\}_{i=1}^N 
   \sim \pi_{\theta_{\text{old}}}(q) \Bigr]  + \frac{1}{N} \sum_{i=1}^N 
   \Bigl\{ \\
&\;  \min[s_1 \cdot A_i, \; s_2 \cdot A_i] 
   - \beta D_{KL}\!\bigl[\pi_\theta \| \pi_{\text{ref}}\bigr] \Bigr\}
\end{aligned}
\end{equation}

\begin{equation}
\begin{aligned}
s_1 &= \frac{\pi_\theta(o_i|q)}{\pi_{\theta_{\text{old}}}(o_i|q)},
&\;
s_2 &= \mathrm{clip}\!\left(
\frac{\pi_\theta(o_i|q)}{\pi_{\theta_{\text{old}}}(o_i|q)}, \;
1-\epsilon, \; 1+\epsilon
\right)
\end{aligned}
\end{equation}

where $\beta$ is a hyperparameter that controls the degree of the KL loss.
Finally, the policy model is optimized using the GRPO objective with KL-divergence regularization.

\section{Experiments}
\subsection{Dataset}
We evaluate our method on the REC-8K dataset \cite{dai2024referring}, which contains 8,011 images annotated with referring expression–count pairs. The dataset is split into 4,923 images (10,555 pairs) for training, 1,566 images (3,336 pairs) for validation, and 1,522 images (3,231 pairs) for testing.

\subsection{Implementation Details}
All experiments are conducted within the VLM-R1 training framework \cite{shen2025vlmr1}. From the REC-8K training set, we randomly sample 2,000 Image–RE pairs to construct the training subset, denoted as Subset-2K. The maximum prompt length is set to 2048 tokens, and the model generates 8 candidate outputs per input. Training is performed with a batch size of 2 per GPU and 2 gradient accumulation steps, resulting in an effective batch size of 32. The model is trained for 4 epochs using \texttt{bf16} precision.

\subsection{Evaluation Metrics}
Following prior work in object counting \cite{zhang2016single}, we evaluate model performance using Mean Absolute Error (MAE) and Root Mean Squared Error (RMSE).

\begin{table}[htbp]
	\centering
	\caption{Comparison of Results on REC dataset.}
	\resizebox{\linewidth}{!}{
	\begin{tabular}{lcrrrr}
		\toprule
		\multicolumn{1}{c}{\multirow{2}[4]{*}{Method}} & \multirow{2}[4]{*}{Dataset} & \multicolumn{2}{c}{Val Set} & \multicolumn{2}{c}{Test Set} \\
		\cmidrule{3-6}          &       & \multicolumn{1}{c}{MAE} & \multicolumn{1}{c}{RMSE} & \multicolumn{1}{c}{MAE} & \multicolumn{1}{c}{RMSE} \\
		\midrule
		\rowcolor[rgb]{ .906,  .902,  .902} \multicolumn{6}{c}{Specific Models} \\
		Ground-D Swin-T\cite{liu2024grounding} & --   &  11.77     &  28.60     &    11.71   & 26.97   \\
		Ground-D Swin-T\cite{liu2024grounding} & all data   &  9.03     &  21.98     &    8.88   & 21.95   \\
		Ground-D REC\cite{dai2024referring} & all data   &  6.80     &  18.13     &    6.50   & 19.79   \\
		\rowcolor[rgb]{ .906,  .902,  .902} \multicolumn{6}{c}{Open-Source Models} \\
		Qwen2.5VL-3B-Instruct\cite{bai2025qwen2} & --   &  9.24     &  22.54     &    9.67   & 27.98   \\
		+ SFT & all data    &  8.63     &  22.66     &    8.90   & 25.04  \\
		+ FGRPR\cite{wang2025crowdvlm} & Subset-2K    &  5.65     &  17.41     &    5.86   & 19.00  \\
		+ \textbf{ours} & Subset-2K   &  \textbf{5.48}     &  \textbf{16.27}     &    \textbf{5.48}   & \textbf{16.51}   \\
		Qwen2.5VL-7B-Instruct\cite{bai2025qwen2} & -    &  9.02     &  22.60     &    9.10   & 28.18   \\
		+ SFT & all data   &  8.27     &  20.64     &    9.05   & 26.23   \\
		+ FGRPR\cite{wang2025crowdvlm} & Subset-2K   &  5.29     &  16.20     &    5.67   & 23.21   \\
		+ \textbf{ours} & Subset-2K   &  \textbf{5.05}     &  \textbf{15.87}     &    \textbf{5.06}   & \textbf{13.92}  \\
		\bottomrule
	\end{tabular}%
	}
	\label{tab:main}%
\end{table}%

\subsection{Main results}
As shown in Table \ref{tab:main}, this study first benchmarked two task-specific architectures, Grounding Dino Swin-T~\cite{liu2024grounding} and Grounding Dino REC~\cite{dai2024referring}, against general-purpose vision-language models. Grounding Dino Swin-T (detection-based) achieved MAEs of 11.77 (val set) and 11.71 (test set) without fine-tuning; using all annotations reduced this MAE to 9.03. Grounding Dino REC (REC-specific) performed better with MAEs of 6.80 and 6.05. Open-source Qwen2.5-VL models remained competitive without task adaptation, with MAEs of 9.24/9.67 (3B) and 9.02/9.10 (7B), highlighting VLMs’ inherent REC generalization.

Subsequently, the impact of different adaptation strategies was investigated. Supervised fine-tuning on the full dataset gave modest improvements; for example, Qwen2.5-VL-7B’s MAE dropped from 9.02/9.10 to 8.27/9.05. Fuzzy Group Relative Policy Reward (FGRPR) \cite{wang2025crowdvlm} was used as a robust baseline. It integrates GRPO with fuzzy rewards to boost fine-grained counting accuracy. Applied to Qwen2.5-VL, FGRPR achieved MAE/RMSE of 5.65/5.86 (3B) and 5.29/5.67 (7B), which was the strongest baseline before our method. However, our approach outperformed both Grounding Dino REC and existing VLM adaptations, setting new state-of-the-art MAEs: 5.48/5.48 (3B) and 5.05/5.06 (7B), while using only 2,000 training samples. 
Our method’s efficacy comes from integrating accuracy and format rewards. The accuracy reward combines range-based and answer-based rewards; the format reward enforces structured responses to improve stability. These designs let VLMs surpass task-specific architectures, validating our reinforcement-guided optimization for REC.

\subsection{Ablation study}
\label{ssec:subhead}
\noindent In our "think-range-answer" paradigm, the range prediction is designed as an auxiliary task to provide coarse spatial-semantic constraints for the primary counting objective. Given its nature as a coarse estimation, the weight of the range reward ($\alpha$) should remain relatively low to prevent it from overshadowing the precision of the final count. To verify this, we evaluated $\alpha \in \{0, 0.2, 0.4, 1.0\}$. 

As shown in Table 2, removing the range reward entirely ($\alpha=0$) results in a validation/test MAE of 5.68/5.62, which already marginally outperforms the baseline (5.65/5.86) due to the non-linear feedback from the Gaussian-guided reward. Conversely, setting $\alpha=1.0$ (Row 3) demonstrates that an excessively high weight for the auxiliary range reward is counterproductive. This causes the policy to disproportionately prioritize interval fitting over final counting precision, leading to a sub-optimal reasoning trajectory as evidenced by the Test RMSE surging to 24.39. The optimal balance is achieved at $\alpha=0.2$ (Row 5), yielding the lowest errors across both sets (5.48/5.48 MAE). These results confirm our hypothesis that a lightweight, auxiliary range signal effectively guides the reasoning process, thereby enhancing the precision of the final count.

\begin{table}[htbp]
	\centering
	\caption{Comparison of range-based reward functions and their effects on training dynamics.}
	\resizebox{\linewidth}{!}{
	\begin{tabular}{ccccccc}
		\toprule
		\multirow{2}[4]{*}{Model} & \multirow{2}[4]{*}{w/o} & \multirow{2}[4]{*}{$\alpha$} & \multicolumn{2}{c}{Val} & \multicolumn{2}{c}{Test} \\
		\cmidrule{4-7}          &       &       & MAE & RMSE & MAE & RMSE \\
		\midrule
		(1) & baseline  & --   & 5.65  & 17.41  & 5.86  & 19.00  \\
        (2) & without  & 0     & 5.68  & 17.19  & 5.62  & 17.06  \\
		(3) & with  & 1     & 5.56  & 16.75  & 5.68  & 24.39  \\
		(4) & with  & 0.4   & 5.51  & 16.52  & 5.54  & 16.66  \\
		(5) & with & \textbf{0.2} & \textbf{5.48 } & \textbf{16.27 } & \textbf{5.48 } & \textbf{16.51 } \\
		\bottomrule
	\end{tabular}%
	}
	\label{tab:ablation}%
\end{table}%

\begin{figure*}[t]
  \centering
  \small
  \begin{tabular}{@{}ccc@{}}
    \includegraphics[width=0.32\linewidth]{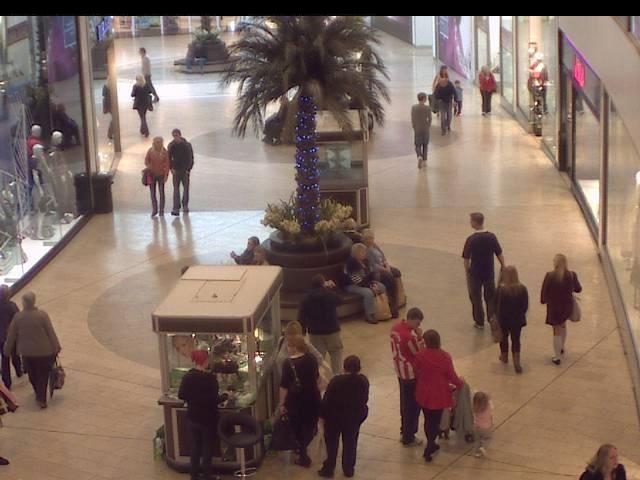} & 
    \includegraphics[width=0.32\linewidth]{figs/5373.jpg} & 
    \includegraphics[width=0.32\linewidth]{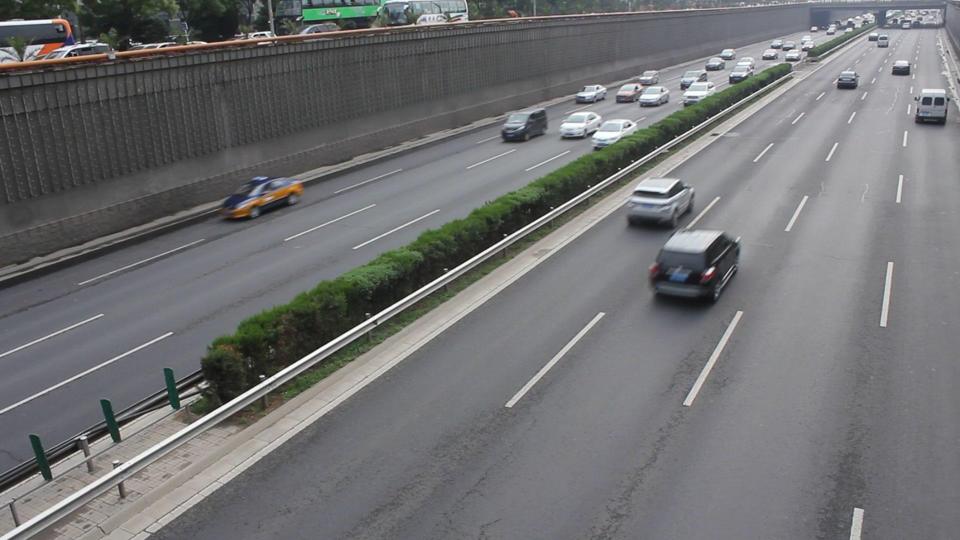} \\
    
    (a) Female person (GT: 19) & (b) Blue glass ball (GT: 14) & (c) Car driving to the left (GT: 24) \\
    \midrule
  \end{tabular}

  \vspace{1mm}
  \begin{tabular}{@{}p{0.3\linewidth}p{0.3\linewidth}p{0.3\linewidth}@{}}
    \textbf{FGRPR:} answer: 17 & \textbf{FGRPR:} answer: 8 & \textbf{FGRPR:} answer: 20\\
    \textbf{Ours:} \textcolor{blue}{range: [15, 20]} \textbf{answer: 19} & 
    \textbf{Ours:} \textcolor{blue}{range: [10, 20]} \textbf{answer: 15} & 
    \textbf{Ours:} \textcolor{blue}{range: [20, 25]} \textbf{answer: 22} \\
  \end{tabular}
  
  \caption{Qualitative comparison. While the baseline provides only a rough estimate, our model predicts a reasonable \textbf{range} (blue) to constrain the reasoning, leading to a final \textbf{answer} more consistent with the Ground Truth (GT).}
  \label{fig:qualitative_results}
\end{figure*}

\subsection{Generalization Analysis}
To further verify the effectiveness of our approach beyond referring expression counting, additional experiments on general object counting were conducted following the experimental setting proposed by FGRPR\cite{wang2025crowdvlm}. Specifically, evaluations were performed on both in-domain categories (e.g., sheep, characters, pedestrians, wheat heads, cars) and the out-of-domain category (manatees), using the dataset and evaluation protocol introduced in FGRPR.
Our method was compared with FGRPR under both 3B and 7B model scales. The results in Table \ref{tab:addlabel} demonstrate that our approach consistently achieves lower MAE across all categories, with distinct improvements in both in-domain and out-of-domain scenarios. These findings confirm that our framework not only performs well in referring expression counting but also exhibits superior generalization on standard counting benchmarks, outperforming previous state-of-the-art methods.

\begin{table}[htbp]
	\centering
	\caption{Cross-domain MAE performance comparison.}
	\resizebox{\linewidth}{!}{%
		\begin{tabular}{c|c|c|c|c|c|c|c} 
			\toprule
			Model & Sheep & Characters & Pedestrians & Wheat Heads & Cars & Manatees & Overall \\ 
			\midrule
			In-VL2-26B\cite{chen2024internvl} & 14.13 & 271.96 & 64.19 & 33.64 & 22.4  & 8.22  & 69.09 \\
			GPT-4o\cite{achiam2023gpt} & 7.12  & 243.41 & 58.36 & 39.18 & 17.22 & 5.82  & 61.95 \\
			LLaMA-90B\cite{dubey2024llama} & 3.87  & 182.4 & 28.6  & 18.18 & 19.24 & 5.7   & 44.68 \\
			Q-3B-SFT & 4.19  & 104.34 & 33.38 & 18.57 & 10.68 & 8.15  & 29.89 \\
			Q-7B-SFT & 4.86  & 97.41 & 24.78 & 18.51 & 11.63 & 7.93  & 27.52 \\
			Q-3B-R1\cite{shen2025vlmr1} & 5.75  & 97.43 & 41.74 & 25.99 & 14.78 & 8.92  & 32.44 \\
			Q-7B-R1\cite{shen2025vlmr1} & 8.04  & 145.55 & 32.29 & 22.68 & 14.42 & 9.88  & 38.81 \\
			Q-3B-FR\cite{wang2025crowdvlm} & 4.19  & 85.52 & 24.02 & 18.86 & 10.55 & 7.81  & 25.16 \\
			Q-7B-FR\cite{wang2025crowdvlm} & 3.33  & 88.1  & 20.6  & 14.52 & 9.71  & 8.16  & 24.07 \\
			\midrule
			\textbf{ours-3B} & \textbf{3.05} & \textbf{83.75} & \textbf{18.74} & \textbf{14.31} & \textbf{9.34} & \textbf{6.64} & \textbf{22.64} \\
			\textbf{ours-7B} & \textbf{2.08} & \textbf{46.15} & \textbf{11.12} & \textbf{12.01} & \textbf{8.63} & \textbf{6.45} & \textbf{14.4} \\
			\bottomrule
		\end{tabular}%
	}
	\label{tab:addlabel}%
\end{table}%

\subsection{Zero-Shot Generalization Analysis}
\label{sec:zero-shot-generalization}
Due to the scarcity of open-source REC datasets, we conducted a zero-shot transfer analysis by adapting the RefCOCO dataset. Since RefCOCO focuses on unique object retrieval, we reformulate it into a binary counting task ($GT \in \{0, 1\}$) by incorporating negative samples, specifically for OOD evaluation without fine-tuning. While this setup is not fully equivalent to original REC tasks, it provides a rigorous test for zero-shot generalization. As shown in Table 4, REC-RL achieves $76.6\%$ accuracy, outperforming the baseline by $5.6\%$. This improvement confirms that our "think-range-answer" paradigm facilitates the transfer of cognitive priors to unseen environments rather than overfitting to specific dataset artifacts. 

\begin{table}[h]
    \centering
    \caption{Out-of-Distribution (OOD) generalization performance on the RefCOCO dataset. The models were tested on a binary existence detection task (500 random samples) without fine-tuning on the target domain.}
    \label{tab:ood_generalization}
    \begin{tabular}{l c c c c}
        \toprule
        \textbf{Method}  & \textbf{MAE} & \textbf{RMSE} & \textbf{Accuracy}\\
        \midrule
        Baseline & 0.29 & 0.54 & 71.0\% \\
        \textbf{Ours} & \textbf{0.23} & \textbf{0.48} & \textbf{76.6\%} \\
        \bottomrule
    \end{tabular}
\end{table}

\subsection{Qualitative Visualization}
\label{sec:visualization}

Figure~\ref{fig:qualitative_results} illustrates the reasoning process of REC-RL in challenging scenarios. While baselines often suffer from hallucinations or miscounts under density and occlusion, REC-RL introduces an intermediate \textit{range} prediction as a cognitive anchor. This mechanism effectively narrows the reasoning search space and facilitates trajectory correction before generating the final count. Coupled with Gaussian-guided rewards for sharper feature alignment, the ``think-range-answer'' paradigm mitigates epistemic uncertainty and ensures robust visual numeracy.

\section{Conclusion}
\label{sec:typestyle}
In this work, we reevaluate the prevailing R1-like training framework for referring expression counting (REC) through the lenses of structural reasoning and non-linear reward shaping. 
First, we introduce the \textit{think-range-answer} paradigm, which reframes REC from a direct mapping task to a structured decision-making process. By treating range prediction as a strategic auxiliary signal rather than a rigid target, this paradigm provides a cognitive anchor that effectively guides visual reasoning and captures quantitative references within complex contexts. Second, we design a multi-component reward system that combines non-linear Gaussian precision guidance with range-based interval supervision. This design allows the model to receive more appropriate feedback regarding the degree of prediction accuracy, which, coupled with GRPO-based optimization, enables the effective acquisition of the proposed reasoning trajectory. Third, comprehensive evaluations on the REC-8K dataset and various OOD benchmarks confirm our method's robust generalization capability. Our ablation studies highlight the critical role of the range-based reward, showing that a lightweight $\alpha = 0.2$ achieves the optimal balance between reasoning constraint and final precision. Ultimately, REC-RL achieves state-of-the-art performance using only 2,000 training samples, proving that a structured reasoning paradigm can significantly enhance data efficiency and model interpretability in multimodal tasks.

\bibliographystyle{IEEEbib}
\bibliography{strings,refs}

\end{document}